\documentclass[runningheads]{llncs}
\usepackage{multirow}
\usepackage[T1]{fontenc}
\usepackage{amsfonts}
\usepackage{graphicx}
\begin{document}
\title{Similarity-based Memory Enhanced Joint Entity and Relation Extraction}
%
%
\author{Witold Kościukiewicz\inst{1,2}\orcidID{0009-0001-0192-8850} \and
Mateusz Wójcik \inst{1,2}\orcidID{0009-0008-0547-9467} \and
Tomasz Kajdanowicz\inst{2}\orcidID{0000-0002-8417-1012} \and
Adam Gonczarek\inst{1}}
\authorrunning{W. Kościukiewicz et al.}
%
\institute{Alphamoon Ltd., Wrocław, Poland \and
Wroclaw University of Science and Technology  \newline
\email{witold.kosciukiewicz@alphamoon.ai}}
\maketitle              
\begin{abstract}
Document-level joint entity and relation extraction is a challenging information extraction problem that requires a unified approach where a single neural network performs four sub-tasks: mention detection, coreference resolution, entity classification, and relation extraction. Existing methods often utilize a sequential multi-task learning approach, in which the arbitral decomposition  causes the current task to depend only on the previous one, missing the possible existence of the more complex relationships between them. In this paper, we present a multi-task learning framework with bidirectional memory-like dependency between tasks to address those drawbacks and perform the joint problem more accurately. Our empirical studies show that the proposed approach outperforms the existing methods and achieves state-of-the-art results on the BioCreative V CDR corpus. 

\keywords{Joint Entity and Relation Extraction \and Document-level Relation Extraction \and Multi-Task Learning}
\end{abstract}
\section{Introduction}
In recent years text-based information extraction tasks such as named entity recognition have become more popular, which is closely related to the growing importance of transformer-based Large Language Models (LLMs). Such models are already used as a part of complex document information extraction pipelines. Even though important pieces of information are extracted, these pipelines still lack the ability to detect connections between them. The missing part, which is the relation classification task, was recognized as a significant challenge in recent years. The problem is even harder to solve if tackled with a multi-task method capable of solving both named entity recognition and relation classification task in a single neural network passage.

\begin{figure}[ht]
\centering
\includegraphics[width=0.7\linewidth]{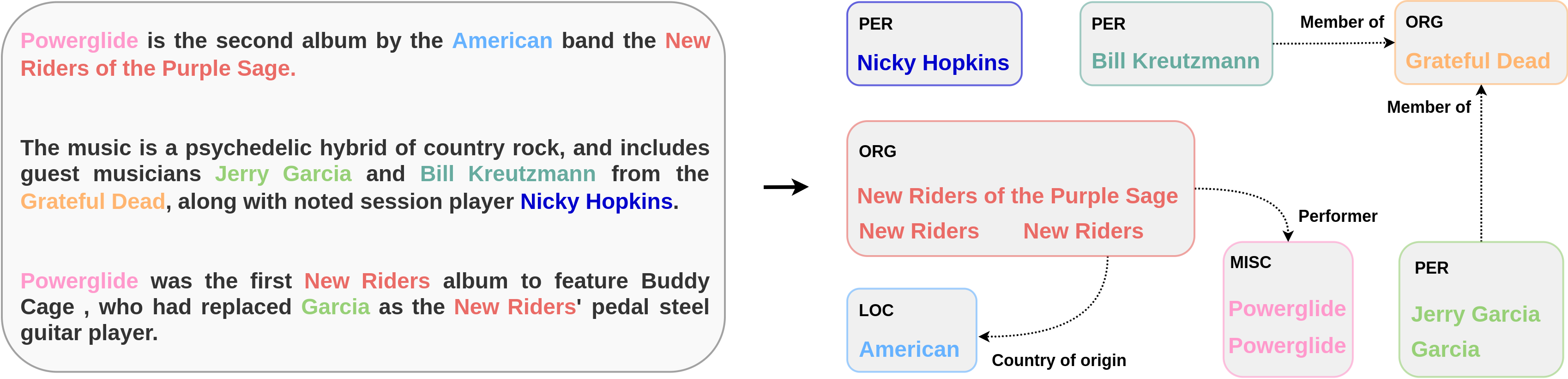}
\caption{\label{fig:problem} Visualization of the document-level joint entity and relation extraction task based on the example taken from DocRED. Entity
mentions originating from different entity clusters are distinguished by color.
}
\end{figure}

In this paper, we propose an approach to solve the multi-task problem of the joint document-level entity and relation extraction problem introduced with the DocRED dataset \cite{yao-etal-2019-docred}. We follow the already existing line of research of learning a single model to solve all four subtasks: mention detection, coreference resolution, entity classification, and relation extraction. The single model is trained to first detect the spans of text that are the entity mentions and group them into coreference clusters. Those entity clusters are then labeled with the correct entity type and linked to each other by relations. Figure \ref{fig:problem} shows an example of a document from the DocRED dataset and a graph of labeled entity clusters that are expected as an output from the model. We introduce the bidirectional memory-like dependency between tasks to address the drawbacks of pipeline-based methods and perform the joint task more accurately.

Our contribution can be summarized as follows: (1) we introduce a new approach that solves multi-task learning problems by improving the architecture of the previously proposed pipeline-based method, introducing the memory module to provide bi-directional dependency between tasks (2) we provide evaluation results, which show that our method outperforms the pipeline-based methods and achieves state-of-the-art results on the BioCreative V CDR corpus (3) we propose a novel similarity classifier module solving distance learning problem for document-level joint entity and relation classification serving as a starting point for future work. The code of our solution is available at \url{ https://github.com/kosciukiewicz/similarity\_based\_memory\_re}.
 
\section{Related work}
\noindent \textit{\textbf{Relation classification}} The relation extraction task is commonly approached by using separately trained models for the Named Entity Recognition \cite{10.1109/TKDE.2020.2981314} to detect entities and then detect relations between them. The transformer-based architectures, pre-trained on large text corpora, such as BERT \cite{devlin-etal-2019-bert}, have dominated the field i.e. Baldini Soares et al. \cite{baldini-soares-etal-2019-matching} uses contextualized input embedding for the relation classification task.

\noindent \textit{\textbf{Joint entity and relation extraction}} The early end-to-end solutions formulated task joint task as a sequence tagging based on BIO/BILOU scheme. These approaches include solving a table-filling problem proposed by Miwa et al. \cite{miwa-sasaki-2014-modeling}. Several approaches tried to leverage multi-task learning abilities using attention-based \cite{katiyar-cardie-2017-going} bi-directional LSTM sharing feature encoders between two tasks to improve overall performance. The inability of the BIO/BILOU-based models to assign more than one tag to a token resulted in using the span-based  method for joint entity and relation extraction proposed in Lee et al. \cite{lee-etal-2017-end}.  Becoming a standard in recent years, this approach was further extended with graph-based methods like DyGIE++ \cite{wadden2019entity} or memory models like TriMF \cite{shen2021trigger}  to enhance token span representation to an end-to-end approach for the joint task.

\noindent \textit{\textbf{Document-level relation extraction }} Although the DocRED \cite{yao-etal-2019-docred} was originally introduced as relation classification benchmark, the opportunity arose to tackle a more complex joint entity and relation extraction pipelines consisting of mention detection, coreference resolution, entity classification, and relation classification.  Since many relations link entities located in different sentences, considering inter-sentence reasoning is crucial to detect all information needed to perform all sub-tasks correctly. Eberts and Ulges \cite{eberts-ulges-2021-end} proposed JEREX - an end-to-end pipeline-based approach showing an advantage in joint training of all tasks rather than training each model separately. In recent work \cite{cabot2021rebel,giorgi2022sequence}, the problem is tackled using a sequence-to-sequence approach that outputs the extracted relation triples consisting of two related entities and relation type as text. 

\section{Approach}

\begin{figure*}[ht]
\centering
\includegraphics[width=0.75\linewidth]{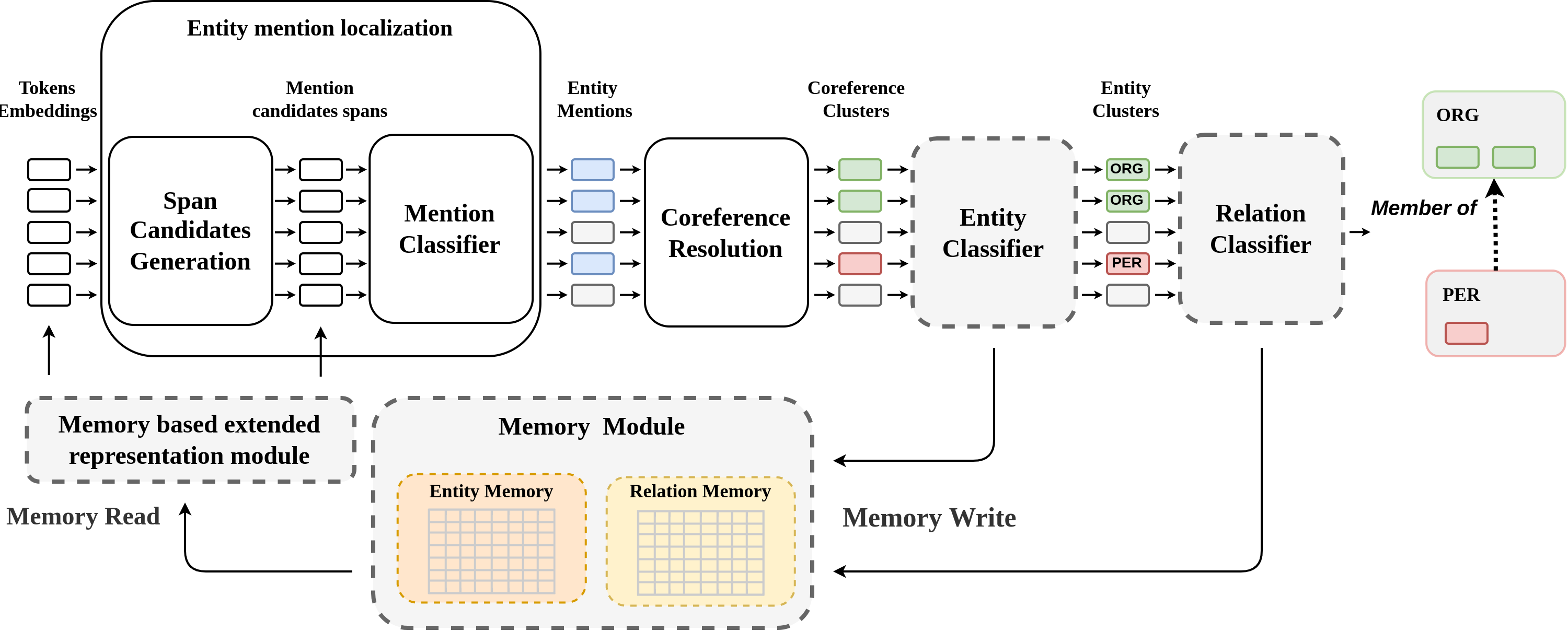}
\caption{\label{fig:model-architecture} The proposed architecture based on JEREX \cite{eberts-ulges-2021-end} enhanced with a feedback loop from entity and relation classifiers to the input of the mentioned classifier step. The novel part of the architecture is highlighted with a gray background and dashed borders. }
\end{figure*}

Our document-level relation extraction framework is inspired by JEREX \cite{eberts-ulges-2021-end} which consists of four task-specific components: mention extraction ($\mathcal{M}$), coreference resolution ($\mathcal{C}$), entity extraction ($\mathcal{E}$) and relation extraction ($\mathcal{R}$). We change the original one-after-another pipeline architecture, introducing the memory module presented in Figure \ref{fig:model-architecture}. The input representations of task-specific models are altered using the memory-based extended representation module that reads the memory using the Memory Read operation. The memory matrices $\mathbf{M_\mathcal{E}}$ and $\mathbf{M_\mathcal{R}}$, are written by the entity and relation classifier, respectively.  That feedback loop allows to share the information with previous steps extending their input by introducing a bi-directional dependency between tasks.

\subsection{Memory reading}

Similarly to TriMF \cite{shen2021trigger}, our approach memory reading is based on the attention mechanism that extends the input representation with the information read from memory. In our architecture, as shown in Figure \ref{fig:model-architecture} we extend both token embeddings $\mathbf{X}_T$ and mention candidates span representations $\mathbf{X}_S$.
For every input representation $\mathbf{X}_i$ where $i\in\{T,S\}$ and memory matrix $\mathbf{M}_j$ where $j\in\{\mathcal{E},\mathcal{R}\}$, the attention mechanism takes the representation $\mathbf{X}_i \in \mathbb{R}^{n \times h}$ as keys and values, where $n$ denotes the number of representations vectors and $h$ is the embedding size. 

As a query, the attention mechanism uses the memory matrix $\mathbf{M}_j \in \mathbb{R}^{m \times s}$ where $m$ denotes the number of memory slots and $s$ is the size of the memory slot. To compute the attention weights vector $\mathbf{a}_{i,j} \in \mathbb{R}^n$ we sum over the memory slots dimension as follows:

\begin{equation}
(\mathbf{a}_{i,j})^\top=\sum_{k}\mathrm{softmax}(\mathbf{m}^{k,:}_j\mathbf{W}^{read}_{i,j}\mathbf{X}^\top_{i})
\end{equation}

\noindent where $\mathbf{W}^{read}_{i,j} \in \mathbb{R}^{s \times h}$ is a learnable parameter matrix for the attention mechanism and $\mathbf{m}^{k,:}_j$ is the $k$-th row of $\mathbf{M}_{j}$. The $\mathbf{a}_{i,j}$ vector is then used to weight the $\mathbf{X}_{i}$ to generate extended input representation $\mathbf{X}'_{i,j}$:

\begin{equation}
\mathbf{X}'_{i,j} = \mathrm{diag}(\mathbf{a}_{i,j})\mathbf{X}_{i}
\end{equation}

 For each input representation $i$, the memory reading operation creates two extended representations $\mathbf{X}'_{i,\mathcal{E}}$ and $\mathbf{X}'_{i,\mathcal{R}}$, based on both memory matrices. The final extended representation is then calculated, using the element-wise mean of $\mathbf{X}_{i}$,  $\mathbf{X}'_{i,\mathcal{E}}$ and $\mathbf{X}'_{i,\mathcal{R}}$:
 
\subsection{Memory writing}

Both memory matrices $\mathbf{M}_\mathcal{E}$ and $\mathbf{M}_\mathcal{R}$ store representations for entity and relation categories respectively. Values encoded in those matrices are written using the gradient of the loss function from the associated classifier -- the entity classifier for $\mathbf{M}_\mathcal{E}$ and the relation classifier for $\mathbf{M}_\mathcal{R}$. To make the stored representations more precise, the loss depends on the similarity between category embedding and the representation of the instance that belongs to that category according to the instance label. As a result, both entity and relation classifiers rely on similarity function $S$ between input representation and suitable memory matrix. The probability distribution over entity types of entity $e_i$ based on its representation vector $\mathbf{x}^e_i$ is calculated as follows:

\begin{equation}
p(\mathbf{y}_e | e_i)=\mathrm{softmax}(S(\mathbf{x}^e_i,\mathbf{M}_{\mathcal{E}}))
\end{equation}

\noindent To get the existence probability over relation types for entity pair $p_{i,j}$ represented by entity pair representation  $\mathbf{x}^{p}_{i,j} \in \mathbb{R}^{h}$ we used the sigmoid function:

\begin{equation}
p(\mathbf{y}_r | p_{i,j})=\mathrm{sigmoid}(S(\mathbf{x}^{p}_{i,j},\mathbf{M}_{\mathcal{R}}))
\end{equation}

\noindent  We define $S$ as bilinear similarity between instance representation $\mathbf{x}$ and memory matrix $\mathbf{M}$ as follows:

\begin{equation}
S(\mathbf{x},\mathbf{M}) = S_{bilinear}(\mathbf{x},\mathbf{M}; \mathbf{W}) = \mathbf{M}\mathbf{W^\top}\mathbf{x}
\end{equation}

\noindent  where $\mathbf{W}$ is a learnable parameter matrix. For both entity and relation classifiers, separate learnable bilinear similarity weight matrices are used: $\mathbf{W}^{write}_{\mathcal{E}} \in\mathbb{R}^{h_e \times s_{\mathcal{E}}}$ and $\mathbf{W}^{write}_{\mathcal{R}} \in \mathbb{R}^ {h_p \times  s_{\mathcal{R}}}$ where $h_e$ and $h_p$ denote entity and entity pair representation sizes respectively. $s_{\mathcal{E}}$ and $s_{\mathcal{R}}$ denote the memory slot size of the entity and relation memory matrices. In our approach number of slots for the memory matrices are equal to the number of types in associated classifiers.

\subsection{Training}

Finally, our model is trained optimizing the joint loss $\mathcal{L}^{joint}$ which contains the same four, sub-tasks related, loss $\mathcal{L}^j$ weighted with fixed, task-related weight value $\beta_j$ as in JEREX \cite{eberts-ulges-2021-end}:

\begin{equation}
\mathcal{L}^{joint} =  \beta_{\mathcal{M}}\mathcal{L}^\mathcal{M} + \beta_{\mathcal{C}}\mathcal{L}^\mathcal{C} + \beta_{\mathcal{E}}\mathcal{L}^\mathcal{E} + \beta_{\mathcal{R}}\mathcal{L}^\mathcal{R}.
\end{equation}

We also include the two-stage training approach proposed in TriMF \cite{shen2021trigger}, tuning the \textit{memory warm-up proportion} during the hyperparameter search.

\section{Experiments}

\noindent \textit{\textbf{Datasets}} We compare the proposed similarity-based memory learning framework to the existing approaches using DocRED \cite{yao-etal-2019-docred} dataset which contains over 5000 human-annotated documents from Wikipedia and Wikidata. By design, DocRED dataset was intended to be used as a relation classification benchmark but its hierarchical annotations are perfectly suitable for joint task evaluation.  For train, dev, and test split we follow the one provided in JEREX \cite{eberts-ulges-2021-end}. According to recent work \cite{tan2022revisiting}, DocRED consists of a significant number of false negative examples. We used dataset splits provided with Re-DocRED \cite{tan2022revisiting} which is a re-annotated version of the DocRED dataset. We also provide results on one area-specific corpus annotated in a similar manner as DocRED - BioCreative V CDR \cite{li2016biocreative} that contains 1500 abstracts from PubMed articles. Following the prior work \cite{christopoulou-etal-2019-connecting,giorgi2022sequence} we used the original train, dev, and test set split provided with the CDR corpus.

\noindent \textit{\textbf{Training}}  As a pretrained text encoder we used $BERT_{BASE}$ \cite{devlin-etal-2019-bert}. For the domain-specific BioCreative V CDR dataset we used $SciBERT_{BASE}$ \cite{Beltagy2019SciBERT} which was trained on scientific papers from Semantic Scholar. All classifiers and memory module parameters were initialized randomly. During training, we used batch size 2, AdamW optimizer with learning rate set to \textit{5e-5} with linear warm-up for 10\% of training steps and linear decay to \textit{0}. The stopping criteria for training were set to 20 epochs for all experiments.

\noindent \textit{\textbf{Evaluation}} During the evaluation we used the strict scenario that assumes the prediction is considered correct only if all subtasks-related predictions are correct. We evaluated our method using micro-averaged F1-score. In Section \ref{section:results} we reported F1-score for a final model evaluated on the test split. As the final model we selected the one that achieved the best F1-score measured on the dev split based on 5 independent runs using different random seeds. Our evaluation technique follows the one proposed in \cite{eberts-ulges-2021-end,giorgi2022sequence}.

\noindent \textit{\textbf{Hyperparameters}} All hyperparameters like embedding sizes or multi-task loss weights were adopted from the original work \cite{eberts-ulges-2021-end} for better direct comparison. Our approach introduces new hyperparameters for which we conducted grid search on the dev split to find the best value. That includes hyperparameters such as \textit{memory warm-up proportion} \cite{shen2021trigger}, memory read gradient, number and types of memory modules, and finally the size of memory slots.

\section{Results}
\label{section:results}

\begin{table*}
\setlength{\tabcolsep}{8pt}
\centering
\caption{\label{tab:main_results} Comparison (F1-score) of our method on the relation extraction task with existing end-to-end systems. $*$ - results from original publications.}
\scalebox{0.85}{
\begin{tabular}{lrrr}
\scriptsize	
 \textbf{Model} &  \textbf{CDR} &  \textbf{DocRED} & \textbf{Re-DocRED} \\
\hline
JEREX \cite{eberts-ulges-2021-end}  &     $42.88$ &   $^*40.38$ &     $\textbf{45.56}$ \\
seq2rel \cite{giorgi2022sequence}  &     $^*40.20$ &   $^*38.20$ &     $-$ \\
\textit{ours}  &     \textbf{43.75} &  \textbf{40.42} &     44.37 \\
\hline
JEREX$_{pre-training}$ &       - &   41.27 &     45.81 \\ $\textit{ours}_{pre-training}$ &       - &   \textbf{41.75} &     \textbf{45.96} \\
\end{tabular}
}
\end{table*}

In Table \ref{tab:main_results} we present a comparison between our approach and existing end-to-end methods on 3 benchmark datasets for joint entity and relation extraction. The provided metric values show that our approach outperforms existing methods on CDR by about $0.9$ percent points (pp.), achieving state-of-the-art results. Our method achieves similar results on DocRED and is outperformed by JEREX architecture on Re-DocRED dataset. We argue that the memory warm-up proportion value (0.4) is too small to properly initialize memories with accurate category representation. On the other hand increasing the memory warm-up steps leaves no time to properly train memory read modules. To address this issue we conducted experiments on pre-trained architecture using distantly annotated corpus of DocRED dataset to initialize memory matrices. We did the same pre-training for JEREX and the results show that our approach outperforms the original architecture by up to $0.48$ pp. on both DocRED-based datasets.

\begin{table}[ht]
\setlength{\tabcolsep}{8pt}
\centering
\caption{\label{tab:baseline_comparison} Comparison (F1-score) between our architecture including memory reading module with JEREX using different relation classifier components - Global (\textit{GRC}) and Multi-Instance (\textit{MRC}). $*$ - results from original publications.}
\scalebox{0.85}{
\begin{tabular}{llrrr}
\scriptsize	
\textbf{Model} &  \textbf{CDR} &  \textbf{DocRED} &  \textbf{Re-DocRED} \\
 \hline
 \textit{GRC} &     &    &      \\
JEREX \cite{eberts-ulges-2021-end} &     $42.04$ &   $^*37.98$ &     $43.46$ \\
\textit{ours} &    $42.04$ &   $37.76$ &     $43.64$ \\
\textit{ours} + memory &      $\textbf{42.18}$ &    $\textbf{39.68}$ &      $\textbf{44.77}$ \\
\hline
 \textit{MRC} &     &    &      \\
JEREX \cite{eberts-ulges-2021-end} &   $42.88$ &   $^*40.38$ &     $\textbf{45.56}$ \\
\textit{ours} &    $43.12$ &    $\textbf{40.68}$ &     $44.93$ \\
\textit{ours} + memory &     $\textbf{43.75}$ &   $40.42$ &     $44.37$ \\
\end{tabular}
}
\end{table}

For the direct comparison with the original architecture we evaluated our memory-enhanced approach with two relation classifiers modules proposed in \cite{eberts-ulges-2021-end}. Results presented in Table \ref{tab:baseline_comparison} show that our method improves the Global Relation Classifier (\textit{GRC}) on every dataset by up to $1.70$ pp. We also tested the performance of our method without the memory module - only with distance-based classifiers. Based on the results in Table \ref{tab:baseline_comparison}, including a memory module with a feedback loop between tasks, in most cases, improved the final results regardless of the \textit{GRC} or \textit{MRC} module.

\section{Conclusions and future work}

In this paper, we proposed a novel approach for multi-task learning for document-level joint entity and relation extraction tasks. By including memory-like extensions creating a feedback loop between the tasks, we addressed the issues present in the previous architectures. Empirical results show the superiority of our method in performance over other document-level relation extraction methods, achieving state-of-the-art results on BioCreative V CDR corpus. One of the possible directions for future work is further development of the memory module by using different memory read vectors for more meaningful input encoding in enhanced representation module or improving the content written to memory by replacing the bi-linear similarity classifier with different distance-based scoring functions or proposing a different method of writing to memory.

\section{Acknowledgements}
The research was conducted under the Implementation Doctorate programme of Polish Ministry of Science and Higher Education and also partially funded by Department of Artificial Intelligence, Wroclaw Tech and by the European Union under the Horizon Europe grant OMINO (grant number 101086321). It was also partially co-funded by the European Regional Development Fund within Measure 1.1. “Enterprise R\&D Projects”, Sub-measure 1.1.1. “Industrial research and development by companies” as part of The Operational Programme Smart Growth 2014-2020, support contract no. POIR.01.01.01-00-0876/20-00.


%
%
%
%
\bibliography{custom}
\bibliographystyle{splncs04}
\end{document}